\documentclass[11pt]{article}

\usepackage[final]{acl}

\usepackage{times}
\usepackage{latexsym}

\usepackage[T1]{fontenc}

\usepackage[utf8]{inputenc}

\usepackage{microtype}

\usepackage{inconsolata}

\usepackage{graphicx}

\usepackage{multirow}

\newcommand{\bhline}{\noalign{\hrule height 1.0pt}}

\title{How Do Language Models Acquire Character-Level Information?}

\author{Soma Sato\hspace{2em}  Ryohei Sasano\\
  Graduate School of Informatics, Nagoya University\\
 \texttt{sato.soma24@gmail.com} \hspace{2em} 
 \texttt{sasano@i.nagoya-u.ac.jp}}

\begin{document}
\maketitle
\begin{abstract}
Language models (LMs) have been reported to implicitly encode character-level information, despite not being explicitly provided during training. 
However, the mechanisms underlying this phenomenon remain largely unexplored. 
To reveal the mechanisms, we analyze how models acquire character-level knowledge by comparing LMs trained under controlled settings, such as specifying the pre-training dataset or tokenizer, with those trained under standard settings.
We categorize the contributing factors into those independent of tokenization. 
Our analysis reveals that merge rules and orthographic constraints constitute primary factors arising from tokenization, whereas semantic associations of substrings and syntactic information function as key factors independent of tokenization. 
\end{abstract}

\section{Introduction}
In recent years, language models (LMs) have made substantial progress in a wide range of natural language processing tasks. 
These models are thought to acquire advanced understanding and generation capabilities through pre-training on massive text corpora, capturing high-level linguistic information, such as lexical, syntactic, and contextual aspects~\cite{Devlin2019BERT, Radford2019GPT2, tenney2019bert}.
Meanwhile, the units used in LMs training are subwords, which are segmented by the tokenizer, and finer-grained information at the character level is not explicitly provided during training.
It might therefore be assumed that LMs do not retain information about the characters within each subword.
Prior studies, however, have shown that information about characters and substrings within subwords is, to some extent, reflected in LMs’ internal representations~\cite{cute, Kaushal2022, itzhak-levy-2022-models}. 
Nevertheless, although these works establish that LMs learn character-level information, research that elucidates the mechanisms of this acquisition remains largely unexplored.

\begin{figure}[t]
\centering
\includegraphics[width=0.92\linewidth]{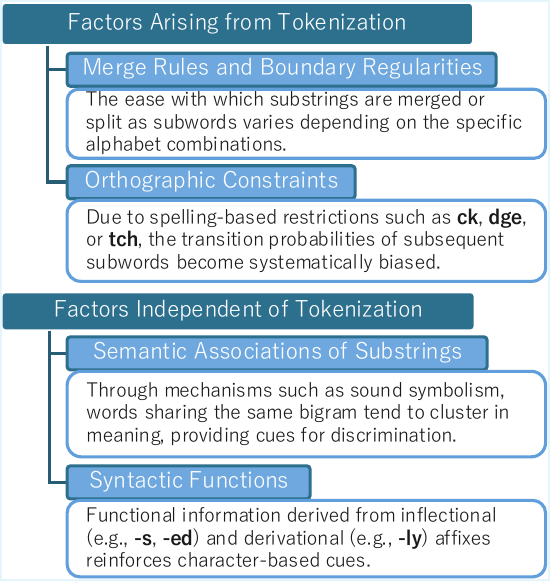}
\vspace{-1ex}
\caption{Factors contributing to character-level information acquisition in LMs.}
\label{fig:intro}
\vspace{-1ex}
\end{figure}

To address this open question, we systematically examine the mechanisms by which LMs acquire character-level information. 
We first manipulate the pretraining data to test whether character-level information can be learned by LMs trained on the manipulated corpora.
To evaluate this, we employ probing classifiers, following \citet{Kaushal2022}, which take token embeddings as input and are trained to make a binary decision about whether a token contains a given alphabetic character.
Building on these results, we categorize the sources of character acquisition into factors arising from tokenization and factors independent of tokenization, as illustrated in Figure~\ref{fig:intro}, and quantify their respective contributions.
Our analysis reveals that merge rules and orthographic constraints constitute primary factors arising from tokenization, while semantic associations of substrings and syntactic information serve as key factors independent of tokenization. 

\section{Background}
\label{section:2}
The possibility that LMs implicitly retain character-level knowledge has been reported through analyses of surface-level information. 
\citet{Kaushal2022} demonstrated through probing experiments on subword embeddings that larger models are more likely to predict the presence of specific characters and substrings.
\citet{itzhak-levy-2022-models} reported that word spellings can be partially reconstructed from LM embedding matrices.
\citet{Hiraoka2024} showed that while exact reconstruction of subword composition is limited, information about token length and substrings is reflected in internal representations.
Furthermore, the CUTE benchmark by \citet{cute} evaluated LMs on tasks probing understanding of string composition and character- and word-level operations, reporting that models are fragile to character-level edits such as insertion and reordering, but exhibit a certain robustness in recognizing inclusion relations. 
Analyses comparing morphologically functional substrings (e.g., prefixes, suffixes and stems) with meaningless substrings also showed that the former tend to carry more substring-level information~\cite{ciaccio-etal-2025-beyond}.

However, these studies stop at showing that LMs encode some information about characters within subwords, and offer only limited analysis of the mechanisms underlying such acquisition. 
In particular, no systematic analysis has been conducted on how individual factors such as merge rules or orthographic constraints in subword segmentation affect character-level information acquisition. 
Our work aims to fill this gap by clarifying the concrete mechanisms of character-level information acquisition through systematic analysis.

We quantitatively analyze how subword segmentation itself affects the learning of character-level information.
As a prerequisite, we briefly review representative subword tokenization algorithms. 
In neural language models, subword representations are widely used to address the out-of-vocabulary (OOV) problem arising from fixed-size vocabularies.
Byte-pair encoding (BPE)~\cite{Sennrich2016bpe} is a greedy tokenization method based on merge rules of frequent character pairs, and has been shown to improve translation accuracy while enhancing vocabulary generalization. 
BPE starts from characters and iteratively merges the most frequent adjacent pairs in the training corpus to obtain a sequence of merge rules, which are then greedily applied for tokenization. 
WordPiece~\cite{Schuster2012WordPiece,Wu2016GNMT} incrementally expands the vocabulary by evaluating the contribution of candidate merges to the increase in corpus log-likelihood, and has been widely adopted in Google’s translation systems and BERT models. 
Unlike BPE, it relies on a likelihood-based criterion for vocabulary updates.  
The unigram language model~\cite{Kudo2018SP,Kudo2018SubwordReg} instead assumes a probabilistic generative model consisting of a subword vocabulary and probabilities for each subword, and optimizes it using the expectation maximization (EM) algorithm~\cite{dempster1977maximum}.
Because segmentation is stochastic, this approach is also known to function as a form of regularization and data augmentation during training.

\begin{figure*}[t]
\centering
\includegraphics[width=0.96\textwidth]{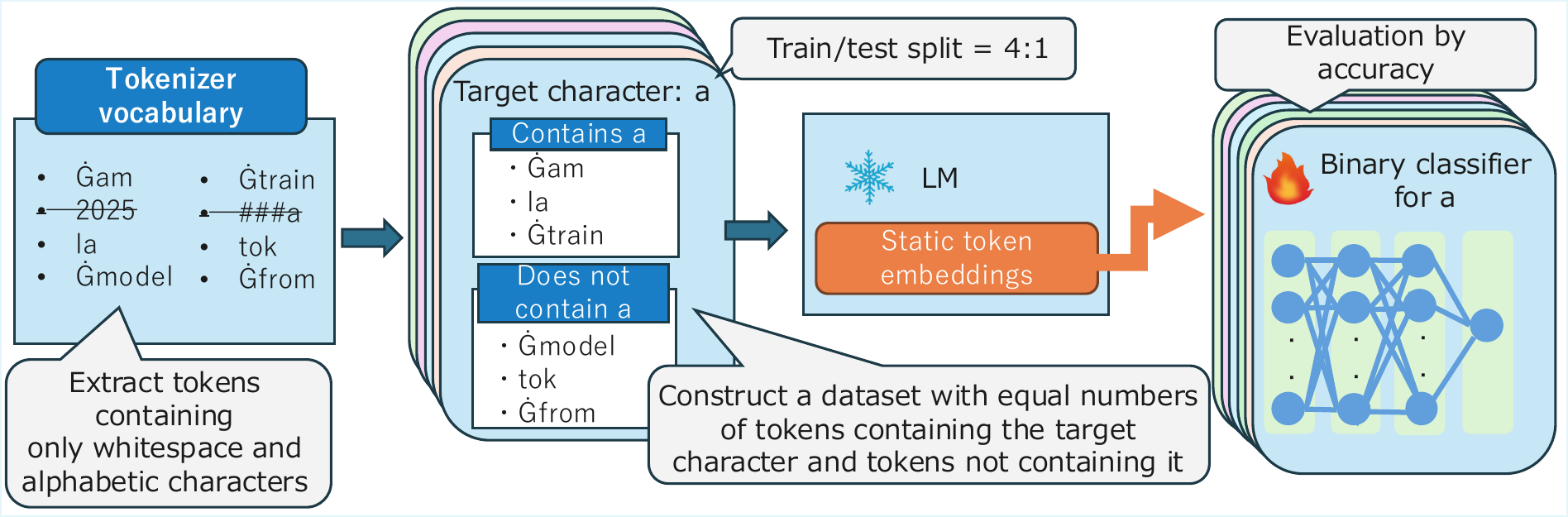}
\caption{Overview of the probing task, which evaluates whether token embeddings encode character-level information in LMs.
\.{G} is a symbol used in BPE tokenization to indicate the space preceding a word.
}
\label{fig:overview}
\end{figure*}

\section{Experimental Setup}
\label{section:3}

In this section, we describe the setup for analyzing the factors underlying character-level information acquisition. 
We first introduce a binary classification task to measure how much character-level information is retained in model representations.
We then pretrain small-scale LMs to verify whether character-level knowledge can indeed be acquired through pretraining.

\subsection{Probing Task}
\label{subsection:3.1}

Following \citet{Kaushal2022}, we quantify the extent to which LMs acquire character-level information by training a multilayer perceptron (MLP) on pretrained token embeddings to predict whether a given token contains a particular alphabetic character. 
However, while longer tokens are more likely to contain a given character, previous work did not account for the effect of token length. 
For example, the probability that a three-character token contains ``e'' is much lower than that of a ten-character token. 
In addition, shorter words tend to be high frequency function words, while longer words are often lower frequency content words, meaning that token length is also correlated with semantic information. 
Thus, pretrained embeddings inherently encode information about token length, which may influence classification performance.
To avoid token length from acting as a confounding factor in classification, we matched the token length distributions between positive and negative examples. 

Figure~\ref{fig:overview} illustrates the overall procedure. 
In this study, we performed tokenization under the constraint that subwords are not allowed to contain whitespace except at the beginning.
We first excluded from the learned tokenizer vocabulary any tokens containing characters other than a leading whitespace symbol (\.{G}) or alphabetic letters.
For each character $\alpha$, we constructed a dataset 
$D(\alpha) = \{(w_1,y_1),\dots,(w_d,y_d)\}$, where $y_i$ indicates whether token $w_i$ contains $\alpha$. 
The numbers of positive and negative examples were balanced.
We used a two hidden layer neural network with SELU activation in the first layer, Tanh in the second layer, and a dropout rate of 0.1. Training was performed with a sigmoid cross entropy loss, the Adam optimizer (learning rate 1e-3), and for 3 epochs.
We report the micro-averaged accuracy across all alphabetic characters.

\subsection{Pretraining Setup}
\label{subsection:3.2}

We verify whether character-level knowledge can be acquired through pretraining. 
Given computational constraints, we adopt two small-scale LMs: BERT-Tiny~\cite{Turc2019BERTTiny} and nanoGPT~\cite{nanoGPT}. 
BERT-Tiny is an encoder-only model with 2 encoder blocks, 2 attention heads, and a hidden size of 128, trained on masked language modeling (MLM). 
In contrast, nanoGPT is a decoder-only model with 12 decoder blocks, 12 attention heads, and a hidden size of 768, trained on next-word prediction.

For the pretraining corpus, we use FineWeb~\cite{Fineweb}, a high quality Web text dataset extracted and filtered from CommonCrawl. 
Specifically, we employ 10B token sample dataset.\footnote{\url{https://huggingface.co/datasets/HuggingFaceFW/fineweb/viewer/sample-10BT}} We first trained tokenizers on this dataset, applying WordPiece for BERT-Tiny and BPE for nanoGPT.\footnote{In this study, the tokenizer is trained such that whitespace characters do not appear at the beginning of subwords.}

For BERT-Tiny, we set the input length to 256 tokens, the batch size per GPU to 64, the number of epochs to 1, the MLM mask rate to 0.15, the learning rate to 2e-5, and the gradient clipping value to 1.0.
For nanoGPT, we used a micro-batch size of 32 with gradient accumulation of 40, which corresponds to an effective batch size of 1,280, and trained with a sequence length of 1,024.
Each iteration processed approximately 1.31M tokens, and we trained for 10,000 iterations, amounting to about 13.1B tokens in total.\footnote{We trained on 10B tokens and then continued training on an additional 3.1B overlapping tokens from the beginning of the dataset.}
The learning rate was 6e-4 with weight decay 1e-1, using 2,000 warm-up iterations followed by cosine annealing scheduling.
On a single A100 GPU, pretraining took 85 hours for BERT-Tiny and 24 hours for nanoGPT.

\subsection{Preliminary Experiments}

We applied our pretrained BERT-Tiny and nanoGPT models to the probing task, and compared them against publicly released models: GPT-2~\cite{Radford2019GPT2}, BERT-base-uncased~\cite{Devlin2019BERT}, and GPT-J~\cite{gpt-j}.
The `matched' column in Table~\ref{tab:1} presents the experimental results.\footnote{\citet{Kaushal2022} report that GPT-J, GPT-2, and BERT-base-uncased achieve higher accuracies in Table 1 of their paper. However, a closer look at Figure 2 in the same paper indicates that these values correspond only to the case where the target character is restricted to z. When all alphabet characters are taken into account, the accuracies decline and become comparable to those observed in the present study.}
While our models exhibit slightly lower accuracies than the large public models, they nonetheless acquire character-level information. Notably, nanoGPT achieves accuracy comparable to BERT-base-uncased, making it a suitable target for our subsequent analyses.
A detailed analysis of how accuracy improves throughout the training process is provided in Appendix~\ref{appendix:A}.
We therefore focus primarily on nanoGPT in the following sections.

\begin{table}[tb] 
\small
\centering
\begin{tabular}{lcc}\bhline
& \multicolumn{2}{c}{\scriptsize \textbf{Length dist.}}\\
& Matched & Unmatched \\\bhline
{\scriptsize \textbf{Public models}} & &  \\
 \ \ GPT-J & 84.0 & 88.0  \\
 \ \ GPT-2 & 75.5 & 79.4  \\
 \ \ BERT-base-uncased & 67.1 & 69.1  \\
\hline
{\scriptsize \textbf{Our trained models}} & &  \\
 \ \ nanoGPT & 66.7 & 69.6  \\
 \ \ BERT-Tiny & 60.8 & 63.3  \\
\bhline
\end{tabular}
\caption{Comparison of publicly available models and our trained models.
`Matched' indicates the accuracy when the token length distribution between positive and negative examples is matched, while `Unmatched' indicates the accuracy without matching.} 
\label{tab:1}
\end{table}

The column labeled unmatched in Table 1 shows the performance in the probing task when the length distributions of positive and negative examples are not aligned. 
This setting, adopted in prior studies, randomly selects positive examples from vocabulary subwords containing the target character and negative examples from those not containing it. 
As a result, a bias arises whereby longer subwords are more likely to be chosen as positives, leading to slightly higher accuracy compared to the matched setting we adopt.

To further examine the degree to which character-level information is learned depending on subword length, we conducted additional experiments with nanoGPT.
We divided the test data into five categories based on subword length: 3 or fewer, 4, 5, 6, and 7 or more. We then measured the classification accuracy for each category.
The proportions of these categories in the overall vocabulary were 10.9\% (3 or fewer), 15.9\% (4), 15.6\% (5), 14.9\% (6), and 42.7\% (7 or more).
We then conducted evaluations using only the data included in each category. 
It should be noted that the original datasets were constructed to match the length distributions between positive and negative examples, ensuring that the numbers of positives and negatives are balanced in every subset.
Table~\ref{tab:1.1} presents the results. 
These results indicate that shorter subwords yield higher probing task accuracy.

\begin{table}[tb] 
\small
\hbox to\hsize{\hfil
\begin{tabular}{l|ccccc|c}\bhline
Length & $\leq 3$ & 4 & 5 & 6 & 7$\leq$ & All \\\bhline
Acc. & 80.4 & 71.9 & 66.7 & 63.9 & 62.3 & 66.7\\
\bhline
\end{tabular}\hfil}
\caption{Accuracy of the probing task across groups defined by token length.} 
\label{tab:1.1}
\end{table}

\section{Analysis via Input String Transformations}
In this section, we analyze how different properties of input strings affect subword embeddings on encoding character-level information.
Specifically, we investigate how applying various transformations to the input text affects performance on the probing task.

\subsection{Design of Input Transformations}
The accuracy observed for nanoGPT may partly reflect correlations between word forms and meanings. 
For example, cognitive effects such as the bouba/kiki phenomenon~\cite{ramachandran2001synaesthesia} suggest that associations between form and meaning can systematically influence learning, and such lexical cues may contribute to character acquisition.
To test this hypothesis, we constructed transformed versions of the FineWeb dataset.

First, to remove the influence of associations between form and meaning, we performed a transformation called word substitution (WordSub), in which each subword in the tokenizer vocabulary was replaced with a uniquely determined random string. 
This preserves grammatical structure and sentence-level semantics, but eliminates systematic relations between form and meaning.

Second, to verify that models do not acquire character-level information when trained on data in which the regularities of character sequences other than whitespace are completely eliminated, we applied a transformation called character perturbation (CharPert), in which each alphabetic character was randomly replaced while preserving letter case information.
For example, given the sentence ``she dreams she is dreaming'', CharPert yields a randomized string such as ``xwe kmdawa dmj ie dkjrumez''. 
In contrast, WordSub consistently replaces \textit{she} with the same substitute (e.g., \textbf{wij}), producing a sentence such as ``\textbf{wij} gkaswd \textbf{wij} da qwmmfans''.
Table~\ref{tab:transform_examples} illustrates concrete examples of these transformations.

\begin{table}[t]
\small
\centering
\begin{tabular}{l|l}
\bhline
Method & Example \\
\bhline
Original & she dreams she is dreaming. \\
WordSub & wij gkaswd wij dj qwdmfans. \\
CharPert & xwe kmdaws dmj ie dkjsrumez.  \\
\bhline
\end{tabular}
\caption{Examples of input string transformations using WordSub and CharPert.}
\label{tab:transform_examples}
\end{table}

In addition, we compared the setting using a BPE tokenizer (BPE) with the setting using words as tokens defined by whitespace segmentation without any tokenizer (Word), in order to examine the effect of tokenization on character-level information learning.
The original vocabulary size was 50,256, and the target vocabulary consisted of 42,875 tokens in the BPE setting and 42,736 tokens in the Word setting.

\subsection{Results}
\label{subsection:4.2}
Table~\ref{tab:2} summarizes the results. 
Although we expected the accuracy for the CharPert dataset to be around 50.0\%, the BPE setting achieved 58.2\%.
This indicates that factors arising purely from subword segmentation algorithm contribute to character-level information acquisition, independent of linguistic properties.  
Next, for the WordSub dataset, accuracies decreased relative to FineWeb by 12.9\% (BPE) and 11.2\% (Word), confirming that associations between form and meaning play a strong role in learning character-level information.
Overall, these results show that both factors arising from tokenization and factors independent of tokenization contribute to character-level information acquisition. 
In the following sections, we analyze these two perspectives in greater detail.

\begin{table}[tb] 
\small
\hbox to\hsize{\hfil
\begin{tabular}{l|ccc}\bhline
& FineWeb & WordSub & CharPert \\\bhline
BPE & 66.7 & 53.8 & 58.2\\
Word & 61.2 & 50.0 & 50.0  \\
\bhline
\end{tabular}\hfil}
\caption{Comparison of accuracy with and without tokenization and the effect of pretraining data differences.} 
\label{tab:2}
\end{table}

\section{Analysis of Factors Arising from Tokenization}
\label{section:5}
We consider two factors arising from tokenization: the merge rules that construct subwords and the orthographic constraints between subwords within words.
In this section, we design tokenizers to isolate these factors and quantitatively assess their impact.

\subsection{Effect of Merge Rules}
\label{subsection:5.1}

\begin{figure}[t]
\centering
\includegraphics[width=\columnwidth]{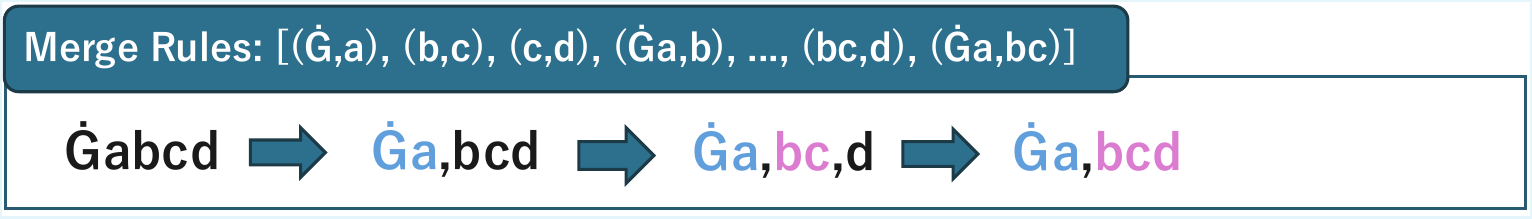}
\caption{Case where merge (b,c) has higher priority than (c,d).}
\label{fig:combine-1}
\vspace{-1ex}
\end{figure}

\begin{figure}[t]
\centering
\includegraphics[width=\columnwidth]{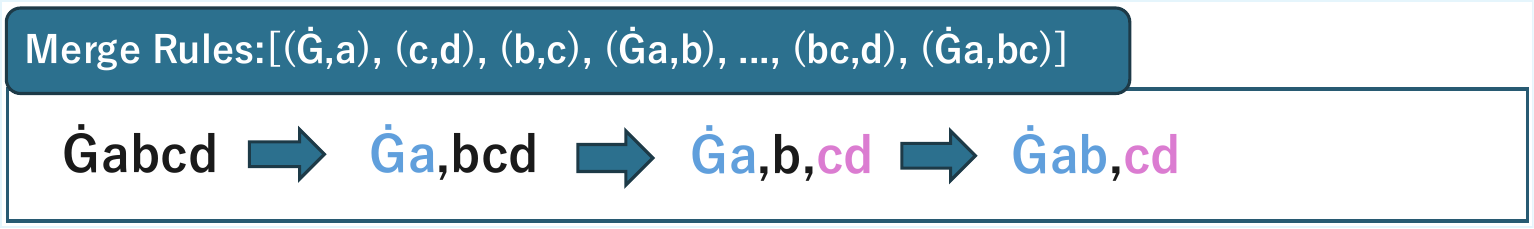}
\caption{Case where merge (c,d) has higher priority than (b,c).}
\label{fig:combine-2}
\vspace{-1ex}
\end{figure}

A key difference between settings with and without subword segmentation on the CharPert dataset is whether information about the segmentation procedure is preserved or not.
Suppose the BPE merge rules are defined as
[(\.{G},a), (b,c), (c,d), (\.{G}a,b), \ldots, (bc,d), (\.{G}a,bc)].
These rules are applied in the listed order of priority.
Consider how the token \.{G}abcd would be segmented. It is first split into characters [\.{G}, a, b, c, d]. Applying the first rule ({\.{G}},a) yields [\.{G}a, b, c, d]. Applying (b,c) gives [\.{G}a, bc, d]. Since neither (c,d) nor (\.{G}a,b) is applicable, (bc,d) applies, resulting in [\.{G}a, bcd].
The flow of the series of processes is shown in Figure~\ref{fig:combine-1}.
If, alternatively, the merge (c,d) had higher priority than (b,c), we would first apply (c,d) to obtain [\.{G}a, b, cd] after [\.{G}a, b, c, d].
We would then apply (\.{G}a,b) to obtain [\.{G}ab, cd], 
The flow of the series of processes is shown in Figure~\ref{fig:combine-2}.

In this way, if the priority of a merge rule is low, the corresponding string is more likely to be split.
As a result, the model parameters statistically accumulate the tendency for a subword ending with a particular character to be followed by a subword beginning with certain characters, which influences the acquisition of character-level information.

\subsubsection{Analysis Procedure}
To evaluate the effect of merge rules on character-level information acquisition by LMs, we use a controlled tokenizer designed under fully specified conditions. 
In BPE, merge rules depend on training data, making it difficult to disentangle data properties from rule effects. 
Explicit definition of the merge rules allows us to directly analyze how they help language models acquire character-level information.

Concretely, we restrict the tokenizer vocabulary to combinations of lowercase alphabetic characters with lengths from 1 to 3, both with and without a leading whitespace symbol.
The resulting vocabulary size is 36,556.\footnote{There are 26, 676 (=$26^2$), and 17,576 (=$26^3$) combinations of lengths 1, 2, and 3, respectively, totaling 18,278. Doubling this to account for the presence/absence of the leading whitespace gives 36,556.}
We configure merge rules so that odd-length strings prefer longer prefixes and even-length strings split evenly.
For example, abc is formed from (ab, c), and \.{G}def from (\.{G}d, ef).
Although the order of merge rules is shuffled, we define their application to proceed in the following sequence: ⟨\.{G}+1 character⟩, ⟨2 characters⟩, ⟨\.{G}+2 characters⟩, ⟨3 characters⟩, and ⟨\.{G}+3 characters⟩.
Here, \.{G}+n characters means that the whitespace symbol \.{G} is followed by n characters, and ⟨⟩ denotes merge rules.

To remove information arising from the merge rules, we additionally construct a pretraining dataset by applying a random token substitution to the tokenized CharPert corpus, preserving whether tokens include a leading whitespace and their length.
For example, a sequence such as ``\.{G}def, ac, \.{G}def, \.{G}qda'' is randomly replaced with ``\.{G}fda, po, \.{G}gfs, \.{G}rer.''
This procedure eliminates statistical patterns induced by the merge rules, allowing us to quantify the contribution of the rules themselves.

Our analysis proceeds as follows. 
We first tokenize CharPert dataset with the controlled tokenizer, and then create a randomly substituted variant of the tokenized data.
We pretrain nanoGPT on each dataset and evaluate both models on the probing task.
To analyze merge rule effects in more detail, we further partition positive examples into six groups according to two factors: (i) whether they include a leading whitespace \.{G}, and (ii) the position of the target character within the subword, i.e., first, second, or third character.
We then report accuracies for each condition.

\begin{table}[bt]
\centering
\small
\begin{tabular}{l|ccc|c}\bhline
& 1st char & 2nd char & 3rd char & all\\\bhline
w/ \.{G} & 49.8 & 54.4 & 90.6 & \multirow{2}{*}{62.4} \\
w/o \.{G} & 73.8 & 61.2 & 72.5\\\bhline
\end{tabular}
\caption{
Probing task accuracy with a controlled tokenizer.
The rightmost column reports the results obtained with all positive examples, whereas the three central columns report results from experiments where the positive examples were partitioned into six groups.}
\label{tab:3}
\end{table}

\subsubsection{Results}
Table~\ref{tab:3} shows the experimental results.
Using the controlled tokenizer on the CharPert dataset, the accuracy reached 62.4\%, exceeding the 50.0\% level even without training a tokenizer on the dataset.
In contrast, when we applied random substitution to the tokenized CharPert data, accuracy dropped to 50.1\%. 
This indicates that, in BPE, merge rules themselves impact character-level information acquisition.
Table~\ref{tab:3} also reports experimental results where the positive examples were partitioned into six groups, defined by the target character position and the presence or absence of the leading whitespace \.{G}.
The results indicate that accuracy differs substantially across conditions.

\begin{figure}[t]
\centering
\includegraphics[width=\columnwidth]{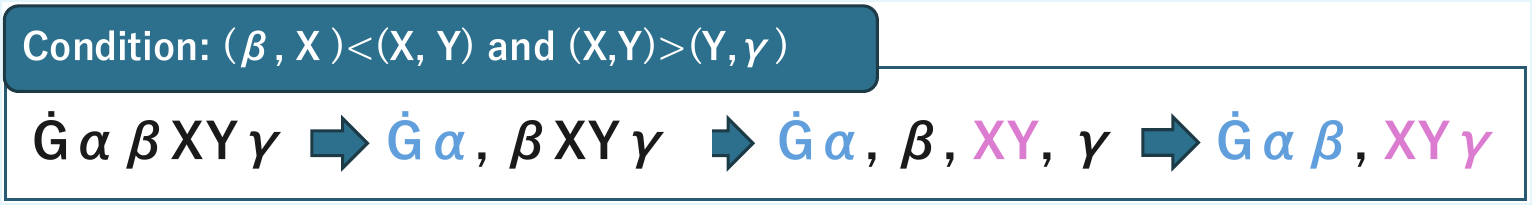}\
\caption{The case where {G}$\alpha\beta$, XY$\gamma$ is segmented as {G}$\alpha\beta$ and XY$\gamma$.}
\label{fig:combine-3}
\vspace{-1ex}
\end{figure}

\begin{figure}[t]
\centering
\includegraphics[width=\columnwidth]{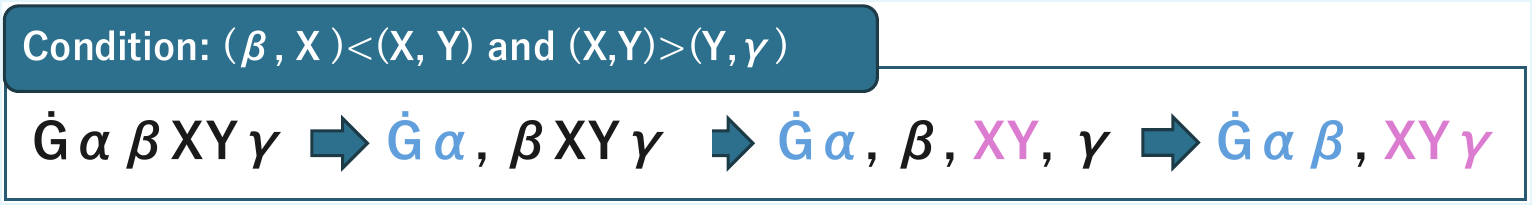}
\caption{The case where {G}$\alpha\beta$X$\gamma\delta$ is segmented as {G}$\alpha\beta$ and X$\gamma\delta$.}
\label{fig:combine-4}
\vspace{-1ex}
\end{figure}

We now consider why the accuracies for the first and second characters with a leading whitespace are high (Figure~\ref{fig:combine-3}).
Suppose \.{G}$\alpha\beta$XY$\gamma$ is segmented as [\.{G}$\alpha\beta$, XY$\gamma$], where X and Y are specific alphabetic characters, and $\alpha$, $\beta$, and $\gamma$ are arbitrary alphabetic characters.
Initially, the sequence is segmented into individual characters: [\.{G}, $\alpha$, $\beta$, X, Y, $\gamma$].
Next, since processing begins with ⟨\.{G}+1 character⟩, the sequence becomes [\.{G}$\alpha$, $\beta$, X, Y, $\gamma$].
For the segmentation [\.{G}$\alpha\beta$, XY$\gamma$] to be obtained, the sequence must first become [\.{G}$\alpha$, $\beta$, XY, $\gamma$]. The necessary condition for this is that m($\beta$, X) < m(X, Y) and m(X, Y) > m(Y, $\gamma$). 
Here, m(X, Y) denotes the strength of the merge between X and Y, which reflects the degree to which the rule is assigned higher priority in the merge rule table.
Once the segmentation [\.{G}$\alpha$, $\beta$, XY, $\gamma$] is reached, ⟨\.{G}+2 characters⟩ and ⟨3 characters⟩ are subsequently applied, yielding [\.{G}$\alpha\beta$, XY$\gamma$].
In other words, the necessary condition for \.{G}$\alpha\beta$, XY$\gamma$ to be obtained is m($\beta$, X) < m(X, Y) and m(X, Y) > m(Y, $\gamma$), and whether \.{G}$\alpha\beta$ appears depends on the choice of XY.
Information about which combinations are more likely is statistically encoded in the model parameters, and the presence of XY in subword tokens is thus indirectly learned.

As for why the accuracy of the first character without a leading whitespace is higher than that of the second character (Figure~\ref{fig:combine-4}), consider the case where \.{G}$\alpha\beta$X$\gamma\delta$ is segmented as [\.{G}$\alpha\beta$, X$\gamma\delta$], where X is a specific alphabetic character and $\alpha$, $\beta$, $\gamma$, and $\delta$ are arbitrary alphabetic characters.
The condition for obtaining [\.{G}$\alpha\beta$, X$\gamma\delta$] is m($\beta$, X) < m(X, $\gamma$) and m(X, $\gamma$) > m($\gamma$, $\delta$), and which \.{G}$\alpha\beta$ appears depends on the choice of X.

The reasons why the third character without a leading whitespace and the third character with a leading whitespace achieve high accuracies are described in Appendix~\ref{appendix:B}.

\begin{figure}[t]
\centering
\includegraphics[width=1\columnwidth]{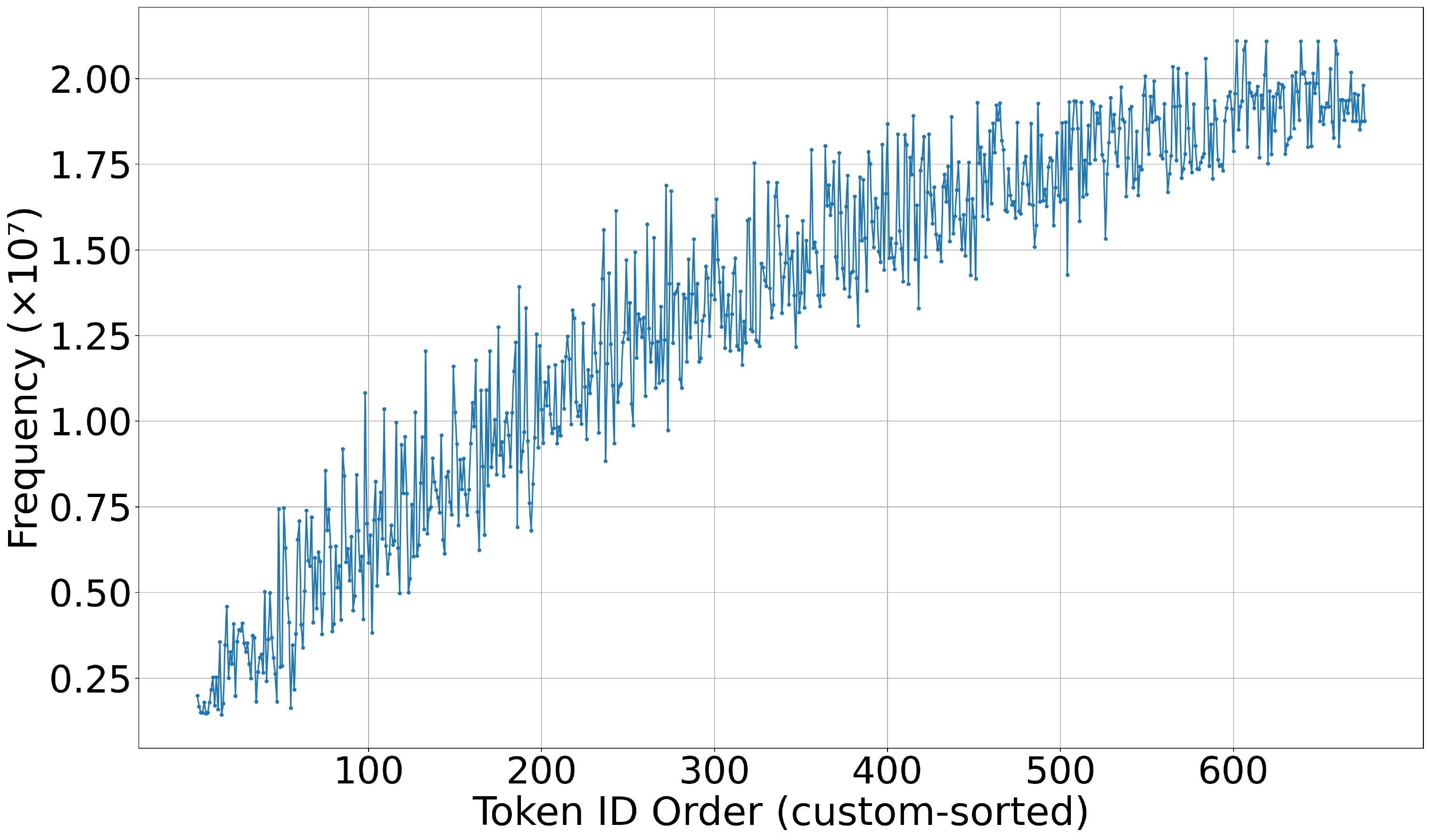}
\caption{Relationship between merge rule strength and the frequency of occurrences at subword boundaries.}
\label{fig:combine}
\end{figure}

\subsubsection{Additional Analysis}
We further analyzed the relationship between the strength of merge rules and the frequency of subword boundaries.
Under controlled merge rules, certain letter combinations are expected to appear more frequently within the same subword, whereas others are more likely to be separated at subword boundaries.
For example, when the letter pair ``ab'' is frequently merged by the rules, many subwords containing ``ab'' are generated, while letter pairs with weaker merge strength tend to be split at subword boundaries.

To test this hypothesis, we aggregated the frequencies of pairs consisting of the final character of each subword and the initial character of the subsequent subword from the tokenized CharPert dataset, and analyzed their correlation with the ordering of merge strengths in the merge rules.

The results are shown in Figure~\ref{fig:combine}.
We observed that subwords with weaker merge strength, that is, those with larger token ID appearing later in the merge rule table, exhibit higher frequencies of the corresponding character pairs occurring at subword boundaries.
These findings empirically support the theoretical explanation that the statistical properties of merge rules directly influence the mechanism by which models acquire character-level information.

\subsection{Effect of Orthographic Constraints}
\label{subsection:5.2}
Natural languages impose orthographic constraints on the adjacency of substrings within words.
English orthography, for example, strongly restricts within word letter sequences. 
Typical patterns include \textsl{ck} rarely appearing word-initially and tending to follow a short vowel (e.g., back, pocket); \textsl{dge} (e.g., badge, hedge) and \textsl{tch} (e.g., catch, kitchen) after short vowels within stems, contrasted with \textsl{ge} and \textsl{ch} elsewhere~\cite{Hayes2006VowelsConsonants}. 
In addition, double consonants (e.g., \textsl{bb} in rabbit) exhibit frequency patterns that depend on the preceding vowel spelling and following letters, reflecting letter-combination effects beyond purely phonological correspondences~\cite{Cassar1997DoubleLetters}.

Hence, subword sequences contain regularities induced by orthographic transition constraints that can be learned during pretraining.
The first subword of a word lacks a preceding dependency, making its internal character content harder to infer. 
On the other hand, for subsequent subwords within a word, orthographic constraints with preceding subwords limit the range of likely patterns, making it possible to infer internal character-level information indirectly from statistical regularities.

\subsubsection{Analysis Procedure}
To isolate orthographic effects, we design a three-character segmentation tokenizer (TCS). 
This tokenizer segments words into units of three characters while preserving words of two characters or fewer as they are, thereby performing segmentation independently of the BPE merge rules.
For example, ``\.{G}enterprise,'' where the leading whitespace is ignored for counting, becomes ``\.{G}ent, erp, ris, e.''

Subwords without a leading whitespace are influenced by orthographic transitions from the immediately preceding subword, whereas subwords with a leading whitespace are not.
We therefore evaluate under conditions that include or exclude a leading whitespace to quantify orthographic effects.
To align the conditions of the BPE tokenizer with those of the TCS, the evaluation data for the probing task was restricted to vocabulary items of at most three characters.
We train both tokenizers and models on FineWeb.

\begin{table}[tb]
\small
\centering
\begin{tabular}{l|cc|c}\bhline
& w/ \.{G}  & w/o \.{G} & Overall \\\bhline
TCS & 50.1 & 80.3 & 68.3\\
BPE & 75.3 & 86.2 & 80.4  \\
\bhline
\end{tabular}
\caption{Probing task accuracy to analyze the effects of orthographic constraints within words, comparing tokenization methods with and without \.{G}.}
\label{tab:4}
\end{table}

\begin{figure*}[t]
\centering
\includegraphics[width=0.95\textwidth]{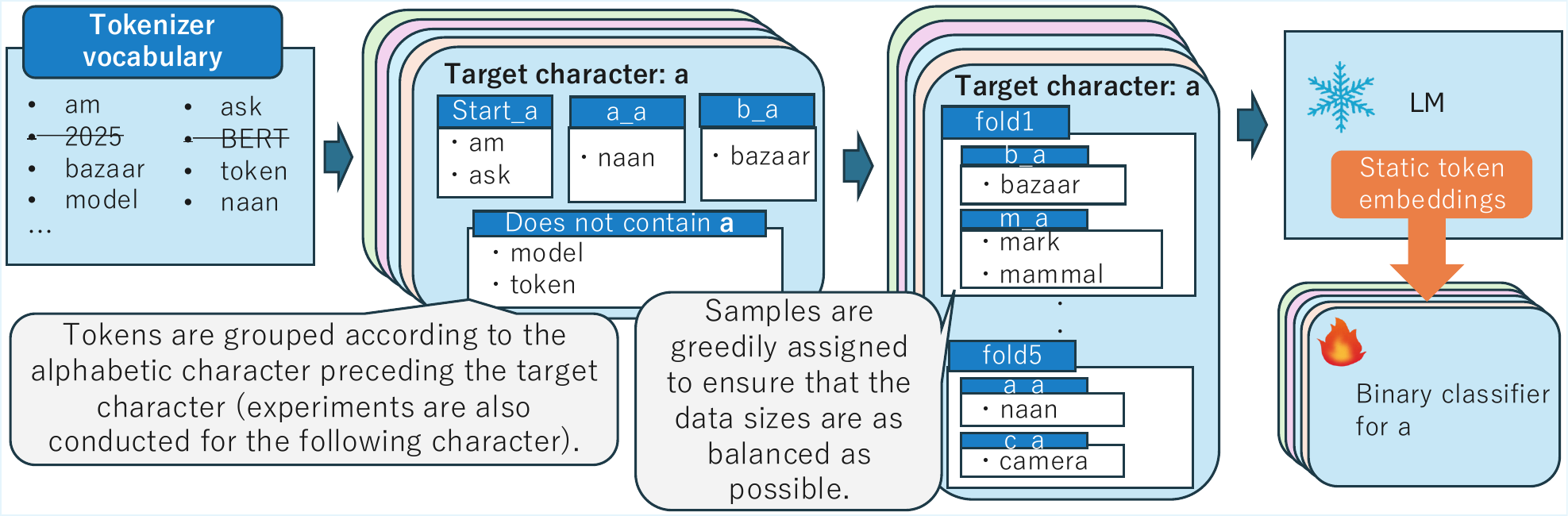}
\vspace{-0.3ex}
\caption{Evaluation Procedure for Semantic Associations of Substrings}
\label{fig:61_img}
\vspace{-0.3ex}
\end{figure*}

\subsubsection{Results}
Table~\ref{tab:4} reports the results. 
Under the whitespace condition with the TCS, the accuracy was 50.1\%, corresponding to the random baseline when the influences of merge rules and orthographic constraints were removed.
In contrast, under the no whitespace condition, the accuracy reached 80.3\%, revealing 30.2 point performance difference attributable to orthographic constraints within words. 
Under the whitespace condition with the BPE tokenizer, the accuracy was 75.3\%, showing 25.2 point performance difference attributable to the effect of merge rules compared with the TCS.
These results indicate that both merge rules and orthographic constraints contribute substantially to character-level acquisition.

\section{Analysis of Factors Independent of Tokenization}
We consider two factors independent of tokenization, namely the semantic associations of substrings and syntactic information.
In this section, we train language models on the FineWeb dataset without subword segmentation and quantitatively analyze their effects on the probing task.

\subsection{Semantic Associations of Substrings}
Sound symbolism can induce semantic relatedness among groups of words that share similar phonological patterns. 
For example, words beginning with \textsl{sl} such as slip, slide, and slope are associated with the concept of sliding, while words beginning with \textsl{gr} such as grab, grasp, and grip relate to grasping.
Such correspondences between spelling and meaning may place words sharing particular character bigrams close to one another in embedding space, thereby providing cues for acquiring character-level information.

\subsubsection{Analysis Procedure}
In the probing task for a target character $\alpha$, we group tokens by their immediate context and partition the data so that training and evaluation splits contain disjoint groups, as illustrated in Figure~\ref{fig:61_img}.
Concretely, we consider two grouping conditions, one based on the character preceding $\alpha$ and the other on the character following $\alpha$, and measure the effect of semantic associations of substrings as the performance difference from a setting without grouping.
In each setting, we partition the data into five groups with approximately equal sizes and perform five-fold cross validation.

\begin{table}[tb]
\small
\centering
\begin{tabular}{@{ }l@{ }|@{ }c@{ }|@{ }c@{ }}
\bhline
 & Accuracy & $\Delta$ from baseline \\\bhline
Baseline & 61.2 & -- \\
Bigram (prev.) & 57.3 & $-3.9$ \\
Bigram (next) & 57.8 & $-3.4$ \\
Lemmatization & 58.2 & $-3.0$ \\
Stemming & 57.3 & $-3.9$ \\
\bhline
\end{tabular}
\caption{Effects of semantic associations and syntactic functions on character inclusion accuracy, showing performance differences from the baseline.}
\label{tab:5}
\end{table}

\subsubsection{Results}
The rows “Bigram (prev.)’’ and “Bigram (next)’’ in Table~\ref{tab:5} report the results for the above grouping conditions. 
Grouping by the preceding character yields 57.3\% accuracy, and grouping by the following character yields 57.8\%, both showing an approximately 3 point decrease relative to the 61.2\% baseline.
These drops suggest that words sharing certain character bigrams cluster in embedding space due to semantic associations of substrings, and that bigram patterns function as cues for predicting the presence of the target character.

\subsection{Effect of Syntactic Functions}
\label{subsubsection:4.2.3}
Syntactic information can link particular characters to lexical or syntactic functions. For example, characters contained in affixes with syntactic functions, such as “-s” marking plural (e.g., books, cats), “-ly” marking adverbs (e.g., quickly, slowly), and “-ed” marking past tense (e.g., walked, played), may facilitate the learning of character-level information through their functional roles.
Prior work has reported that such characters with syntactic functions exhibit higher classification performance~\cite{Kaushal2022}.
This suggests that inflectional and derivational information are reflected in token embeddings, thereby indirectly providing cues for character-level information.

\subsubsection{Analysis Procedure}
To analyze the impact of syntactic information, we apply lemmatization and stemming to the pretraining corpus, thereby removing inflectional morphemes through lemmatization and both inflectional and derivational morphemes through stemming, and train models on the processed data.
Lemmatization removes inflectional morphemes (e.g., -s: plural, -ed: past, -er: comparative), while stemming further removes derivational morphemes (e.g., -ly: adverbial, -ful: adjectival, -tion: nominal).
We quantify the contribution of syntactic cues by comparing performance before and after these removals. 
We use NLTK’s WordNetLemmatizer and the Porter stemmer~\cite{Miller1995WordNet,Porter1980SuffixStripping}.
Following Section~\ref{subsection:3.2}, we train a BPE tokenizer and nanoGPT on each processed FineWeb variant and evaluate on the probing task.

\subsubsection{Results}
The rows “Lemmatization’’ and “Stemming’’ in Table~\ref{tab:5} summarize the results. Lemmatization yields 58.2\% accuracy, 3.0 point decrease relative to the baseline, indicating that inflectional morphemes provide important cues for character acquisition. Stemming yields 57.3\% accuracy, 3.9 point decrease, further showing that derivational morphemes also contribute.

\section{Conclusion}
\label{section:6}
In this study, we analyzed the factors contributing to character-level information acquisition by applying controlled transformations to input text, and divided these factors into those related to tokenization and those independent of it.
Our experiments demonstrated that subword segmentation contributes through merge rules and orthographic constraints within words, while factors independent of segmentation include semantic associations of substring and syntactic information.
The findings of this study suggest that LMs do not treat words as opaque tokens, but rather indirectly learn character-level patterns, which are part of their internal structure, from multiple different cues.
This could help explain the spelling capabilities and the ability to generate text that leverages the sound of characters that LMs occasionally exhibit.
Future work includes a more detailed analysis of the contribution of each factor identified in this study and applying these insights to develop model architectures and pre-training methods that can more robustly process character-level information.

\section*{Limitations}
In this study, we have identified the main factors underlying the acquisition of character-level information; however, some aspects remain unresolved.
One limitation of this work is that we have not fully controlled for the possibility that the pretraining data themselves contain lexical information.
For example, the models may acquire character-level information from explicit lexical resources such as word lists or spelling rules that specify which alphabetic sequences constitute which words.
Although we attempted to disentangle the factors through input transformations and tokenizer manipulations, the contribution of such lexical information has not been completely eliminated.
Furthermore, the nanoGPT model used in this study is relatively small, and it remains unclear what cues contribute to character-level acquisition in larger models. 
In addition, prior work has reported that character-level information is more easily extracted from low frequency tokens and that rare characters tend to yield high classification performance; however, these aspects were not sufficiently examined in our analysis. 
Another limitation is that our experiments were conducted exclusively on English text.
It remains an open question whether the mechanisms identified in this study generalize to other languages with different orthographic systems, morphological complexity, or writing scripts.
Additionally, we used only FineWeb as the pretraining corpus, which may limit the generalizability of our findings.
Different pretraining corpora with varying characteristics in terms of domain, genre, or text quality could potentially yield different patterns of character-level information acquisition.

\section*{Acknowledgments}
This work was partly supported by JSPS KAKENHI Grant Numbers 24H00727.


\bibliography{custom}

@inproceedings{cute,
  title     = {{CUTE}: Measuring LLMs’ Understanding of Their Tokens},
  author    = {Edman, Lukas and Schmid, Helmut and Fraser, Alexander},
  booktitle = {Proceedings of the 2024 Conference on Empirical Methods in Natural Language Processing (EMNLP 2024)},
  year      = {2024},
  pages     = {3017--3026},
}

@inproceedings{Kaushal2022,
  title     = {What do tokens know about their characters and how do they know it?},
  author    = {Kaushal, Ayush and Mahowald, Kyle},
  booktitle = {Proceedings of the 2022 Conference of the North American Chapter of the Association for Computational Linguistics: Human Language Technologies (NAACL 2022)},
  year      = {2022},
  pages     = {2487--2507},
}

@article{Hiraoka2024,
  title   = {Knowledge of Pretrained Language Models on Surface Information of Tokens},
  author  = {Hiraoka, Tatsuya and Okazaki, Naoaki},
  journal = {arXiv:2402.09808},
  year    = {2024},
}

@inproceedings{Sennrich2016bpe,
  title     = {Neural Machine Translation of Rare Words with Subword Units},
  author    = {Sennrich, Rico and Haddow, Barry and Birch, Alexandra},
  booktitle = {Proceedings of the 54th Annual Meeting of the Association for Computational Linguistics (ACL 2016)},
  pages     = {1715--1725},
  year      = {2016},
}

@inproceedings{Schuster2012WordPiece,
  title     = {Japanese and Korean Voice Search},
  author    = {Schuster, Mike and Nakajima, Kaisuke},
  booktitle = {2012 IEEE International Conference on Acoustics, Speech and Signal Processing (ICASSP 2012)},
  pages     = {5149--5152},
  year      = {2012},
}

@article{Wu2016GNMT,
  title   = {Google's Neural Machine Translation System: Bridging the Gap between Human and Machine Translation},
  author  = {Wu, Yonghui and Schuster, Mike and Chen, Zhifeng and Le, Quoc V. and Norouzi, Mohammad and Macherey, Wolfgang and Krikun, Maxim and Cao, Yuan and Gao, Qin and Macherey, Klaus and Klingner, Jeff and Shah, Apurva and Johnson, Melvin and Liu, Xiaobing and Kaiser, {\L}ukasz and Gouws, Stephan and Kato, Yoshikiyo and Kudo, Taku and Kazawa, Hideto and Stevens, Keith and Kurian, George and Patil, Nishant and Wang, Wei and Young, Cliff and Smith, Jason and Riesa, Jason and Rudnick, Alex and Vinyals, Oriol and Corrado, Greg and Hughes, Macduff and Dean, Jeffrey},
  journal = {arXiv preprint arXiv:1609.08144},
  year    = {2016},
}

@inproceedings{Kudo2018SP,
  title     = {{SentencePiece}: A simple and language independent subword tokenizer and detokenizer for Neural Text Processing},
  author    = {Kudo, Taku and Richardson, John},
  booktitle = {Proceedings of the 2018 Conference on Empirical Methods in Natural Language Processing: System Demonstrations (EMNLP 2018)},
  pages     = {66--71},
  year      = {2018},
}

@inproceedings{Kudo2018SubwordReg,
  title     = {Subword Regularization: Improving Neural Network Translation Models with Multiple Subword Candidates},
  author    = {Kudo, Taku},
  booktitle = {Proceedings of the 56th Annual Meeting of the Association for Computational Linguistics (ACL 2018), Volume 1: Long Papers},
  pages     = {66--75},
  year      = {2018},
}

@inproceedings{Turc2019BERTTiny,
  title     = {Well-Read Students Learn Better: On the Importance of Pre-training Compact Models},
  author    = {Turc, Iulia and Chang, Ming-Wei and Lee, Kenton and Toutanova, Kristina},
  booktitle = {arXiv preprint arXiv:1908.08962},
  year      = {2019},
}

@article{Radford2019GPT2,
  title   = {Language Models are Unsupervised Multitask Learners},
  author  = {Radford, Alec and Wu, Jeffrey and Child, Rewon and Luan, David and Amodei, Dario and Sutskever, Ilya},
  journal = {OpenAI Technical Report},
  year    = {2019},
}

@article{Devlin2019BERT,
  title     = {{BERT}: Pre-training of Deep Bidirectional Transformers for Language Understanding},
  author    = {Devlin, Jacob and Chang, Ming-Wei and Lee, Kenton and Toutanova, Kristina},
  journal   = {Proceedings of the 2019 Conference of the North American Chapter of the Association for Computational Linguistics: Human Language Technologies (ACL 2019)},
  year      = {2019},
  pages     = {4171--4186},
}

@inproceedings{tenney2019bert,
  title     = {{BERT} Rediscovers the Classical {NLP} Pipeline},
  author    = {Tenney, Ian and Das, Dipanjan and Pavlick, Ellie},
  booktitle = {Proceedings of the 57th Annual Meeting of the Association for Computational Linguistics (ACL 2019)},
  year      = {2019},
  pages     = {4593--4601},
}

@inproceedings{itzhak-levy-2022-models,
  title     = {Models In a Spelling Bee: Language Models Implicitly Learn the Character Composition of Tokens},
  author    = {Itzhak, Itay and Levy, Omer},
  booktitle = {Proceedings of the 2022 Conference of the North American Chapter of the Association for Computational Linguistics (NAACL 2022)},
  year      = {2022},
  pages     = {5061--5068},
}

@inproceedings{ciaccio-etal-2025-beyond,
  title     = {Beyond the Spelling Miracle: Investigating Substring Awareness in Character-Blind Language Models},
  author    = {Ciaccio, Cristiano and Sartor, Marta and Miaschi, Alessio and Dell’Orletta, Felice},
  booktitle = {Proceedings of the 63rd Annual Meeting of the Association for Computational Linguistics (ACL 2025)},
  year      = {2025},
  pages     = {11361--11372},
}

@article{Fineweb,
  title   = {The {FineWeb} Datasets: Decanting the Web for the Finest Text Data at Scale},
  author  = {Penedo, Guilherme and Kydlíček, Hynek and Ben Allal, Loubna and Lozhkov, Anton and Mitchell, Margaret and Raffel, Colin and Von Werra, Leandro and Wolf, Thomas},
  journal = {arXiv:2406.17557},
  year    = {2024},
}

@article{ramachandran2001synaesthesia,
  title   = {{Synaesthesia—A} window into perception, thought and language},
  author  = {Ramachandran, Vilayanur S. and Hubbard, Edward M.},
  journal = {Nature Reviews Neuroscience},
  pages   = {43--53},
  year    = {2001},
}

@article{Hayes2006VowelsConsonants,
  author  = {Hayes, Heather and Treiman, Rebecca and Kessler, Brett},
  title   = {Children use vowels to help them spell consonants},
  journal = {Journal of Experimental Child Psychology},
  pages   = {27--42},
  year    = {2006},
}

@article{Cassar1997DoubleLetters,
  author  = {Cassar, Marie and Treiman, Rebecca},
  title   = {The beginnings of orthographic knowledge: Children's knowledge of double letters in words},
  journal = {Journal of Educational Psychology},
  pages   = {631--644},
  year    = {1997},
}

@inproceedings{Miller1995WordNet,
  author    = {Miller, George A.},
  title     = {{WordNet}: A Lexical Database for English},
  booktitle = {Speech and Natural Language: Proceedings of a Workshop Held at Harriman},
  year      = {1992},
}

@article{Porter1980SuffixStripping,
  author  = {Porter, Martin F.},
  title   = {An Algorithm for Suffix Stripping},
  journal = {Program-electronic Library and Information Systems},
  year    = {1980},
}

@article{dempster1977maximum,
  title     = {Maximum Likelihood from Incomplete Data via the {EM} Algorithm},
  author    = {Dempster, A. P. and Laird, N. M. and Rubin, D. B.},
  journal   = {Journal of the Royal Statistical Society: Series B (Methodological)},
  volume    = {39},
  number    = {1},
  pages     = {1--38},
  year      = {1977},
  publisher = {Wiley}
}

@misc{nanoGPT,
  author = {Andrej Karpathy},
  title = {nanoGPT: The simplest, fastest repository for training/finetuning medium-sized GPTs},
  howpublished = {\url{https://github.com/karpathy/nanoGPT}},
  year = {2022}
}

@misc{gpt-j, 
    author = {Wang, Ben and Komatsuzaki, Aran}, 
    title = {{GPT-J-6B: A 6 Billion Parameter Autoregressive Language Model}}, 
    howpublished = {\url{https://github.com/kingoflolz/mesh-transformer-jax}}, 
    year = 2021, 
    month = May 
}

\appendix

\section{Relationship between Training Data Size and Accuracy}
\label{appendix:A}

\begin{figure}[t]
\centering
\includegraphics[width=\columnwidth]{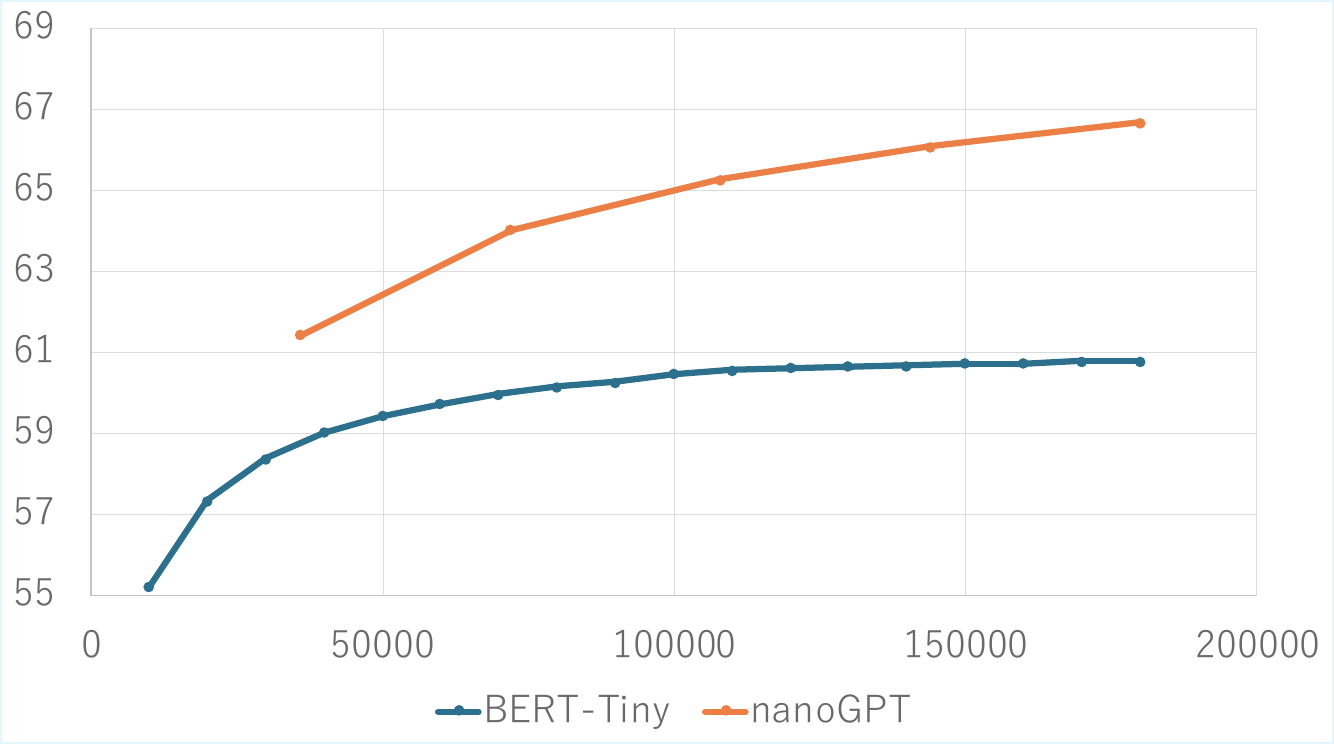}
\vspace{-3ex}
\caption{Relationship between pretraining data volume and accuracy on the character inclusion classification task.}
\label{fig:datasize-acc}
\vspace{-0.3ex}
\end{figure}

In general, it is known that increases in model size and training data volume lead to improved performance on downstream tasks.
In the character inclusion classification task as well, prior research and the results from Section~\ref{section:3} confirm that larger model sizes yield higher accuracy.
On the other hand, the relationship between pretraining data volume and character-level information acquisition is not self-evident.
Therefore, we measure accuracy at the intermediate checkpoints from Section~\ref{subsection:3.2} (18 checkpoints for BERT-Tiny and 5 for nanoGPT) and analyze the relationship with data volume.
The results are shown in Figure~\ref{fig:datasize-acc}.
BERT-Tiny steadily improved from 55.2\% at the first checkpoint to 60.8\% at the final checkpoint, with the growth slowing in the latter half of training.
nanoGPT showed a similar trend, rising from 61.4\% at the first checkpoint to 66.7\% at the final checkpoint.
These results indicate that character-level information is acquired from the early stages of training, and its accuracy further improves as the training data volume increases.

\begin{figure}[t]
\centering
\includegraphics[width=\columnwidth]{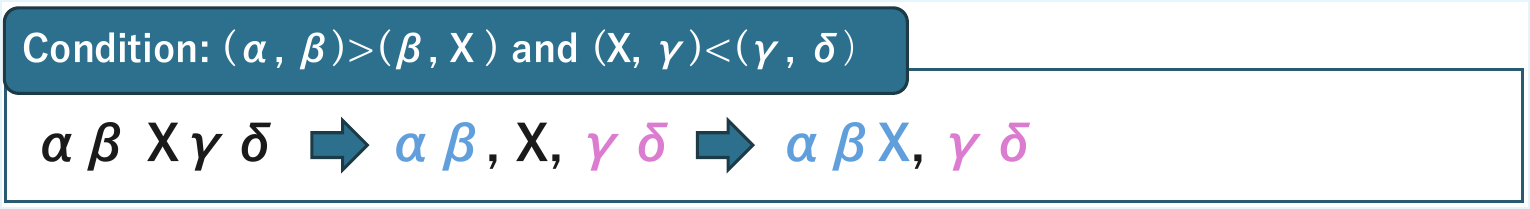}\
\caption{The case where $\alpha\beta$X$\gamma\delta$ is segmented as $\alpha\beta$X and $\gamma\delta$.}
\label{fig:combine-5}
\vspace{-1ex}
\end{figure}

\begin{figure}[t]
\centering
\includegraphics[width=\columnwidth]{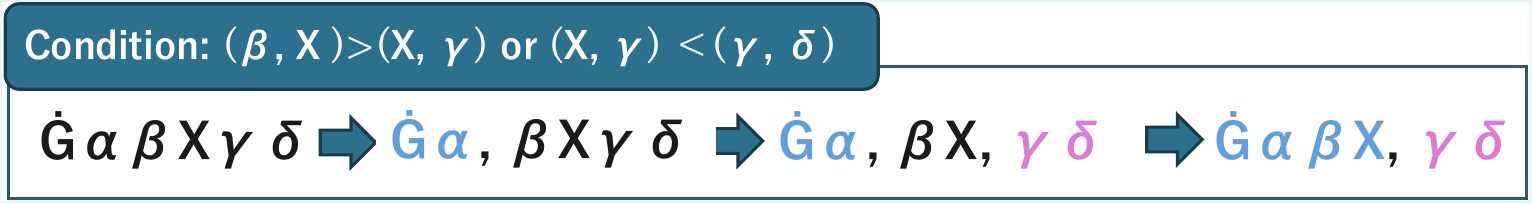}
\caption{The case where {G}$\alpha\beta$X$\gamma\delta$ is segmented as {G}$\alpha\beta$X and $\gamma\delta$.}
\label{fig:combine-6}
\vspace{-1ex}
\end{figure}

\section{Considerations on Merge Rules}
\label{appendix:B}
\paragraph{Why the Accuracy of the Third Character without \.{G} is High}
Consider the case where $\alpha\beta X\gamma\delta$ is segmented as [$\alpha\beta X$, $\gamma\delta$], and Figure~\ref{fig:combine-5} illustrates this sequence of merge,  
where X is a specific alphabetic character, and $\alpha$, $\beta$, $\gamma$, and $\delta$ are arbitrary alphabetic characters.  
First, the sequence is split into [$\alpha$, $\beta$, X, $\gamma$, $\delta$].  
Next, when the ⟨2-character⟩ merge is applied, in order for the sequence to become [$\alpha\beta X$, $\gamma\delta$], it must first take the form [$\alpha\beta$, X, $\gamma\delta$].  
The necessary conditions for this are m($\alpha$, $\beta$) > m($\beta$, X) and m(X, $\gamma$) < m($\gamma$, $\delta$).  
After the segmentation [$\alpha\beta$, X, $\gamma\delta$] is obtained, the ⟨3-character⟩ merge is applied, yielding [$\alpha\beta X$, $\gamma\delta$].  
In other words, the necessary condition for [$\alpha\beta X$, $\gamma\delta$] to be obtained is m($\alpha$, $\beta$) > m($\beta$, X) and m(X, $\gamma$) < m($\gamma$, $\delta$), and whether $\gamma\delta$ appears depends on the choice of X.  
Information about which combinations are more likely is statistically encoded in the model parameters, and whether X is included in subwords is indirectly learned.  

\paragraph{Why the Accuracy of the Third Character with \.{G} is High}
Consider the case where \.{G}$\alpha\beta X\gamma\delta$ is segmented as [\.{G}$\alpha\beta X$, $\gamma\delta$], and Figure~\ref{fig:combine-6} shows this sequence of merges,  
where X is a specific alphabetic character, and $\alpha$, $\beta$, $\gamma$, and $\delta$ are arbitrary alphabetic characters.  
First, the sequence is split into [\.{G}, $\alpha$, $\beta$, X, $\gamma$, $\delta$].  
Next, since merging begins with ⟨\.{G}+1 character⟩, the sequence becomes [\.{G}$\alpha$, $\beta$, X, $\gamma$, $\delta$].  
For the segmentation [\.{G}$\alpha\beta X$, $\gamma\delta$] to be obtained, it must take the intermediate form [\.{G}$\alpha$, $\beta$X, $\gamma\delta$].  
The necessary condition for this is m($\beta$, X) > m(X, $\gamma$) or m(X, $\gamma$) < m($\gamma$, $\delta$).  
After [\.{G}$\alpha$, $\beta$X, $\gamma\delta$] is merged, the ⟨\.{G}+3 character⟩ merge is applied, yielding [\.{G}$\alpha\beta X$, $\gamma\delta$].  
In other words, the necessary condition for [\.{G}$\alpha\beta X$, $\gamma\delta$] to be obtained is m($\beta$, X) > m(X, $\gamma$) or m(X, $\gamma$) < m($\gamma$, $\delta$).  
Conversely, when m(X, $\gamma$) is stronger than either m($\beta$, X) or m($\gamma$, $\delta$), the segmentation [\.{G}$\alpha\beta X$, $\gamma\delta$] does not occur.  
Thus, when [\.{G}$\alpha\beta X$, $\gamma\delta$] segmentation does occur, there exist characters $\gamma$ that appear only rarely depending on X, and the model learns character-level information based on these tendencies.  

\end{document}